# Continuous Control for Automated Lane Change Behavior Based on Deep Deterministic Policy Gradient Algorithm

Pin Wang[1]*, Hanhan Li[1], Ching-Yao Chan[1]

*Abstract* — Lane change is a challenging task which requires delicate actions to ensure safety and comfort. Some recent studies have attempted to solve the lane-change control problem with Reinforcement Learning (RL), yet the action is confined to discrete action space. To overcome this limitation, we formulate the lane change behavior with continuous action in a model-free dynamic driving environment based on Deep Deterministic Policy Gradient (DDPG). The reward function, which is critical for learning the optimal policy, is defined by control values, position deviation status, and maneuvering time to provide the RL agent informative signals. The RL agent is trained from scratch without resorting to any prior knowledge of the environment and vehicle dynamics since they are not easy to obtain. Seven models under different hyperparameter settings are compared. A video showing the learning progress of the driving behavior is available[2]. It demonstrates the RL vehicle agent initially runs out of road boundary frequently, but eventually has managed to smoothly and stably change to the target lane with a success rate of 100% under diverse driving situations in simulation.

## I. INTRODUCTION

In recent years there has been a big increase in the application of Reinforcement Learning in a variety of problems such as video games [1], robotics [2], and autonomous vehicles [3].

In the family of RL algorithms, there are two basic divisions, model-based RL and model-free RL. Model-based RL either knows or explicitly learns a model to describe state transition and reward evaluation, which significantly improves sample efficiency. However, such a model is intractable to obtain in practice, especially in autonomous driving scenarios where driving environment is dynamic and complicated. Model-free RL algorithms do not rely on a specific model and learns a good policy directly through trial and error. Therefore, we resort to model-free RL methods in our study.

The application of RL to the control field, in terms of action space, falls into two categories. One is discrete action space where a set of discretized actions are stored as control commands. The other is continuous action space in which any real number can be sampled within the allowed threshold. The use of discrete action space eases the difficulty of finding an optimal action in a finite space, but it scarifies the control accuracy. Driving is a task that requires delicate actions to ensure safety and comfort. To learn realistic driving behaviors, we formulate the problem in continuous action space.

Regarding the use of training samples, RL has on-policy and off-policy algorithms. On-policy algorithms estimate the behavior policy while using it for generating sample trajectories, while off-policy algorithms use a policy different from the policy being learnt, to collect training samples. For better sample efficiency and low variance, we resort to off-policy methods for learning driving behaviors.

Different RL algorithms, targeting different aspects of the aforementioned properties, have been proposed for continuous control problems. For example, Guided Policy Search [4] is a model-based method that uses locally optimized trajectory to guide the training. Trust region-based policy optimization method [5] is a policy gradient method that guarantees monotonic improvement of the policy at each update. Value based methods (e.g. Deep Deterministic Policy Gradient, DDPG [6]) can use off-policy techniques and model-free properties to estimates deterministic policies with stochastic exploration.

Based on the analysis of the autonomous driving problems in continuous control space, we apply DDPG for learning driving behaviors, particularly in the challenging case of lane change behavior. Importantly, we do not leverage any prior knowledge of the environment and vehicle kinematics and train the RL agent from scratch through a well-designed reward function.

It is worth noting that the RL control policy we learned with DDPG is a mid-level control module, which takes in mid-level representations from the sensing and perception system and outputs mid-level control commands for the low-level controllers.

The rest of the paper is organized as follows. A literature review of related work is given in Section II. Section III introduces the methodology used in our work. Section IV depicts model training and validation in simulated environment. Concluding remarks and discussions of future research directions are provided in the last section.

## II. RELATED WORK

RL has been applied to decision making and driving control problems. For decision making, the action space is typically a set of discretized decision choices. You, et al. [7] adopted Q-learning method for vehicle decision making in highway driving scenarios, where the agent learns to accelerate, brake, overtake and make turns. Mukadam, et al. [8] employed deep reinforcement learning, with Q-mask technique, to learn a high-level policy for tactical decision making which they broke down into five actions, including no action, accelerate, decelerate, left lane change and right lane change.

For driving control problems, a number of studies treated the control action space as discrete to simplify the problem or improve learning efficiency. Zhang, et al. [9] used Double Q-learning to learn vehicle speed control where the control actions are acceleration, deceleration, and maintaining. Jaritz, et al. [10] applied Asynchronous Actor Critic (A3C) method to learn car control in rally games in an end-to-end framework in which they broke the control commands into 32 classes. Hoel, et al. [11] used Deep Q-network (DQN) to address both speed control and decision making for lane change scenarios.

There are some studies that treat the vehicle control problem in continuous space. Wang, et al. [12] proposed a quadratic Q-function approximator to find optimal actions by solving a quadratic problem. Prior knowledge of the vehicle control mechanism was also used for assisting learning. Liang, et al. [13] presented a Controllable Imitative Reinforcement Learning (CIRL) approach to limit the action exploration in a controllable action space by exploiting the knowledge learned from demonstrations of human experts. The RL algorithm was based on DDPG and applied to the navigation task. Kaushik, et al. [14] also used DDPG to learn overtaking maneuvers in continuous action space in a way of curriculum learning that made the agent first learn simple tasks (lane keeping) and then moved on to complex tasks (overtaking), with the goal of fast overtaking car in front of the RL vehicle. Sallab, et al. [15] compared the effect of using discrete action space and continuous action space for the lane keeping task, with the use of deep Q-network and Deep Deterministic Actor Critic (DDAC) respectively, and concluded that both methods could achieve successful lane keeping behavior but DDAC showed better performance with smoothed actions.

In many of these studies, the reward function is defined in a simplified and limited fashion. In [11], as the authors mentioned, they used a simple reward function which gave positive rewards in normal traveling based on the distance traveled in a time interval and negative reward of a constant (-10) when the vehicle crashed. In [13], the authors designed a reward module in which four reward terms were used but three of them were simply defined as two-value piecewise functions and the other was a constant. Simple reward function can clearly indicate the goodness of an action but it may slow down the agent's learning ability in continuous control problems as it fails to give informative signal for each state-action pair. The same concern lies in the use of a sparse reward. As mentioned in [16], sparse reward needs hard exploration and it is highly unlikely to be successful via random exploration.

To allow efficient and effective learning, a good reward function should provide informative signals for every state and/or action status to reveal useful information in the learning procedure. It is not hard to design a reward function which gives evaluation values, but it is difficult to devise an "enlightening" reward function which can guide the RL agent to move toward the right direction.

In our study, we apply DDPG to learn the continuous control policy of the lane change behavior, particularly the lateral movement, from scratch in an interactive driving environment with a well-define informative reward function.

### III. METHODOLOGY

The DDPG algorithm is a policy-gradient and actor-critic method. In this section we will describe how DDPG is applied to our target problem of the lateral control, and illustrate the capability of our DDPG framework in solving realistic driving problems.

*A. Preliminaries*

*1) Actor-critic algorithm*

In the actor-critic algorithm, the actor, i.e. policy function $\mu(s|\theta^\mu)$, generates an action given the current state. The critic evaluates an action-value function $Q(s,a|\theta^Q)$ based on the output from the actor, as well as the current state. The TD (Temporal-Difference) errors produced from the critic drive the learning in both actor network and critic network.

*2) DDPG*

The core of DDPG is that it uses a stochastic way for the exploration of good actions but estimates a deterministic behavior policy (as in (1)). By doing this it only needs to integrate over state space which makes it much easier to learn the policy, but it also has limitations that it may not explore the full state and action space. To overcome the limitation, we introduce a noise process $N$, a truncated normal, in our study for random exploration.

$$a_t = \mu(s_t|\theta^\mu) \quad (1)$$
$$a_t = \mu(s_t|\theta^\mu) + N_t \quad (2)$$

The actor and critic are designed with neural networks. The deterministic policy gradient theorem [17] provides the update rule for the actor network, while the critic network is updated from the gradients obtained from TD errors, which can be formulated in (3).

$$\nabla_{\theta_\mu}\mu \approx \mathbb{E}_\mu[\nabla_a Q(s,a|\theta^Q)|_{s=s_t, a=\mu(s_t)} \nabla_{\theta_\mu}\mu(s|\theta^\mu)|_{s=s_t}]$$
(3)

As shown in (3), to get the expectation we just need the gradients of the critic network w.r.t. actions, and the gradients of the actor network w.r.t. its parameters.

There are also some practical tricks that are used in DDPG, e.g. the use of experience replay to break temporal correlations in experience tuples and the use of target network for better convergence.

*3) Weights update*

For a mini-batch, the critic network is updated by minimizing the loss in (4). The actor policy is updated with sampled policy gradients as in (5).

$$L = \frac{1}{N}\sum_i(y_i - Q(s_i, a_i|\theta^Q))^2 \quad (4)$$

where $y_i = r_i + \gamma Q'(s_{i+1}, \mu'(s_{i+1}|\theta^{\mu'})|\theta^{Q'}$.

$$\nabla_{\theta_\mu}\mu \approx \frac{1}{N}\sum_i[\nabla_a Q(s,a|\theta^Q)|_{s=s_i,a=\mu(s_i)}\nabla_{\theta_\mu}\mu(s|\theta^\mu)|_{s=s_i}] \quad (5)$$

The updates of the target critic and target actor networks are as in (6) (7).

$$\theta^{Q'} \leftarrow \tau\theta^Q + (1-\tau)\theta^{Q'} \quad (6)$$
$$\theta^{\mu'} \leftarrow \tau\theta^\mu + (1-\tau)\theta^{\mu'} \quad (7)$$

where $\tau$ is an update parameter and can be set as $\tau \ll 1$.

### B. Mid-Level Lateral Controller

What we aim to learn is a mid-level lateral controller which takes mid-level input, representation of the driving environment from well-developed perception systems, and generates mid-level output, representation of control commands to be executed by low-level control modules. The perception module processes raw sensor data and provides the most relevant information of the driving scenarios. The low-level controller converts the mid-level output to values that the actuators will take. We suppose these two modules can work collaboratively with our proposed mid-level controller and they are not covered in this study.

*1) Hierarchical system structure*

The lateral control variable in our study is the derivative of yaw rate, which we call yaw acceleration $a_{lat}$. Such a design is based on the idea of jerk, the derivative of acceleration, so as to ensure the change curve in yaw rate to be continuous and the curve of yaw angle to be smooth.

The driving environment contains a great deal of information, some of which may not directly related to the lane-change task. Therefore, we only extract the most relevant state variables (e.g. ego vehicle's dynamics and surrounding vehicles' dynamics) to be included in the state representation. For the lateral movement, the most useful information is the motion of vehicles on the target lane (e.g. speed, position) and the ego vehicle's own state (e.g. speed, position, orientation, acceleration), as well as the road geometry information (e.g. road curvature).

In the longitudinal direction, we use a well-developed Intelligent Driver Model (IDM) with some adaptation as the longitudinal controller which takes the vehicle dynamics of the ego vehicle and its leader (e.g. relative distance and speed) as input and outputs an acceleration

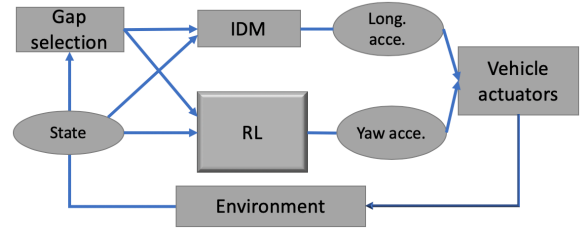

Figure 1. Vehicle system structure

for the ego vehicle to let it follow its leader in desired distance.

In addition, we also resort to a gap selection module to find the proper gap for the vehicle to merge into. In this sense, the gap selection and longitudinal control take charge of the longitudinal movement which, on one hand, ensures driving safety (e.g. no crash with other vehicles) and, on the other hand, relieves the burden of the RL algorithm. The overall system structure in our study is shown in Fig. 1.

The benefit of the use of a RL based lateral control model is that no trajectory plan module is needed for either the ego vehicle or the other surrounding vehicles, and it can handle high dimension input when decision making is incorporated which will be deferred as our future work.

*2) Neural network structure*

In our study, the actor network and critic network are both designed with multi-layer perceptron (MLP). As shown in Fig. 2, The actor takes a state (mid-level representation) as input and outputs an action (mid-level representation), which, in turn, together with the state is fed into the critic network. State-action values, specifically Q-values $Q(s,a|\theta^Q)$ and target Q-values $Q(s,a|\theta^{Q'})$, are generated from the critic network with two different sets of weights, i.e. the (current) critic weights and target critic weights. Loss is calculated based on the state-action values as in (4) and used to update the weights of the critic. The Q value is also a component in the policy gradient procedure of the actor network.

Considering that changes from action and environment between two consecutive time steps may not be quite distinct, we explore the impact of the action update step $k$ on learning performance. In other words, the action value the RL vehicle executes in the driving environment at every time step may be updated with the actor output either at every time step or at multiple time steps. This is to see whether the agent is sensitive enough to notice the change in actions and environment.

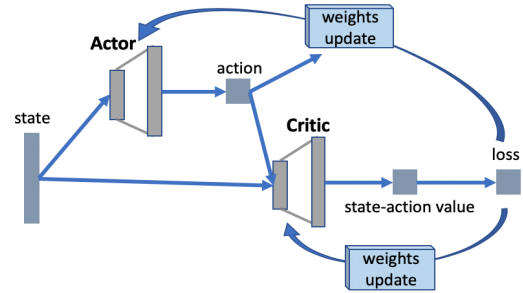

Figure 2. Neural network structure

*3) Reward function*

One big challenge in this study is to learn the continuous control behavior from scratch with no prior knowledge being applied. Therefore, the design of the reward function plays a crucial role in guiding the agent to learn preferred driving policies.

Considering that driving is a delicate task where a small change in action may lead to wrong driving directions or a sequence of small deviations may lead to cascading failures (e.g. deviating from path) and that lane changing is a time sensitive task which needs to be completed as soon as possible once it is initiated, we design the reward function by considering the action values, deviation status and maneuvering time.

In a lane-change process, a smooth transition from one lane to the target lane is reflected in the lateral dynamic control variables. Therefore, we punish large yaw rate, $\omega_{yaw}$, and yaw acceleration, $a_{lat}$, as in (8).

$$r_s = -w_1|\omega_{yaw}| - w_2|a_{lat}| \quad (8)$$

where $w_1$ and $w_2$ are weights, used for adjusting the reward from the two terms to be in similar magnitude.

This part alone may encourage overly defensively behaviors that the vehicle changes lane slowly with small action values. To remedy this, we add another cost term, the step interval $dt$, to evaluate the time duration of the lane-changing maneuver, as in (9). The longer the lane-change process is, the larger the cumulated time will be and so as the penalty. $w_3$ is the weight.

$$r_e = -w_3 dt \quad (9)$$

Another part in the reward is the evaluation of the position deviation, specifically the lateral deviation from the centerline of the target lane, $\Delta d_{lat}$. The lateral deviation reveals the state of the lane-change process. The smaller it is, the closer it is to the target centerline. It is expressed in (10).

$$r_d = -w_4|\Delta d_{lat}| \quad (10)$$

where $w_4$ is the weight.

This part is an important component in the reward function because it is the lateral deviation that directly signifies the success or failure of the lane change behavior. That is, it should get much larger penalty if it goes beyond a certain range of the lane boundary, i.e. border threshold $D$. Furthermore, the penalty when it goes beyond the border threshold should be larger than the reward accumulated when it is lingering around the border but within the threshold. This way, the vehicle agent can get the signal that staying within the boundary is better than going beyond the boundary, which gives guidance in its learning process. Therefore, the reward from lateral deviation is redesigned as in (11).

$$r_d = \begin{cases} -w_4|\Delta d_{lat}|, & |\Delta d_{lat}| \leq D \\ -w_4|\Delta d_{lat}| * 10/(1-\gamma), & |\Delta d_{lat}| \geq D \end{cases} \quad (11)$$

where $D$ is the threshold for lateral deviation and $\gamma$ is the discount factor used action value calculation in (4). (11) means when the deviation is larger than $D$, the penalty it gets will be around 10 times larger than all the cumulated rewards when it lingers around the boarder.

A comprehensive immediate reward for a given state-action pair will be $r = r_s + r_e + r_d$.

IV. SIMULATION AND RESULT

*A. Simulation Environment*

We train the vehicle agent in a simulation environment where a highway segment of 600 meters long is constructed with three lanes on each direction. Vehicles enter the highway with different initial speeds (0-40km/h) and can reach different speed limits (80km/h-120km/h). All the vehicles, except for the RL vehicle agent, travel on its current lane and follow the IDM car-following model, but they can interact with the ego vehicle such as yield or overpass the ego vehicle when it initializes the change lane maneuver, which enriches the experimenting situations.

In each episode, the RL ego vehicle is a randomly selected vehicle on the middle lane. A command of making either right or left lane change will be issued to it after it travels around 80 meters after entrance. After the gap selection module finds a gap, the RL vehicle agent will try to merge to the target lane.

We set two termination conditions for the episode. One is the deviation threshold $D$ which equals to the lane width $W$. That means if the vehicle deviates from the centerline of the target lane over a lane width, it will be terminated. The other is the maneuvering time. If the vehicle is in the lane-changing condition too long ($\geq$10s), the episode will also be terminated, and then a new episode starts again.

*B. Training Settings*

For the actor and critic networks, we use the same neural network structure that comprises two hidden layers, with 300 neurons in the first layer and 600 neurons in the second layer. The learning rate of the actor and critic are set as 0.0001 and 0.001, respectively. The discount factor $\gamma$ is set as 0.98, and the time step interval $dt$ is 0.1s.

The weights in the reward function are set as $w_1 = 10.0$, $w_2 = 5.0$, $w_3 = 0.075$, and $w_4 = 1.0$, with the purpose of making the three rewards terms in similar magnitude.

We found that the replay memory size $D$ used in the critic learning procedure has significant impacts on the learning performance, therefore, we tested three sizes of replay memory: 2000, 4000, 8000.

Additionally, as mentioned earlier, we would like to test the effects of different action update step $k$ and set it as 1, 2, or 4, that is, action value executed in simulation is update at every step, every other step, or every 4 steps while the experience collection and weights update in training procedure are still conducted at every time step.

In total, there are 9 models based on the two hyperparameter combinations, among which we compare 7 models, i.e. in groups of (①, ②, ③), (①, ④, ⑤), (②, ⑥, ⑦), and (④, ⑦), as one control variable varies.

Table 1. Models under different hyperparameters

| k \ D | 2000 | 4000 | 8000 |
|---|---|---|---|
| 1 | ① | ④ | ⑤ |
| 2 | ② | ⑥ | ⑦ |
| 4 | ③ | ⑧ | ⑨ |

## C. Results

### 1) Training results of different models

We first compare the training results on total rewards under different models. In RL, the total reward is the actual return accumulated along the trajectory and can be used to evaluate the performance of the policy. The higher the total reward is, the better the policy is.

Fig. 3(a) shows the impact of the action update step in model ①, ②, ③ where $k = 1, 2, 4$ and $D = 2000$. Model ① (red curve) and ② (blue curve) show better converging trend than model ③, indicating that smaller step size is better.

We then compare the impact of different replay memory size under the two action update steps ($k = 1, 2$). In Fig. 3(b), among models of ①, ④, ⑤ where $D$ is 2000, 4000, 8000 and $k = 1$, model ④ converges fast and has the best performance. In Fig. 3(c), among models of ②, ⑥, ⑦, where $D$ is 2000, 4000, 8000 and $k = 2$, model ⑦ shows the best performance. The results indicate that reasonably large replay memory size is better, but the step size also interrelates with the replay memory size. Hence, we compare model ④ and ⑦ to find the best combination settings. As shown in Fig. 3(d), the dark green curve (model ④) converges faster and is slightly better than light green curve (model ⑦), indicating that model ④ the combination of $k = 1$ and $D = 4000$ is the best choice for the problem.

To further explore the performance of model ④ and ⑦, we compare the accumulated maneuvering steps (i.e. steps while being in a lane change process) over training episodes. It is expected that at the beginning part of the training, the RL vehicle agent does not know how to make lane changes so that the maneuvering steps should be small since it always fails and the lane change is then terminated. As the training goes on, the maneuvering step should grow and come to a steady trend, implying that the RL agent can always make lane changes in stable time windows. Fig. 4(a) shows exactly the expected trend. Model ④ (the dark green curve) demonstrates the expected variation trend earlier than model ⑦ (light green curve), indicating it converges faster than ⑦. This is also proved in Fig. 4(b) where the training loss curve of model ④ shows a more stable converging trend than model ⑦. Therefore, we choose model ④ ($f_a = 1$, $D = 4000$) as the RL model, and use it to evaluate the driving performance discussed in the following subsections.

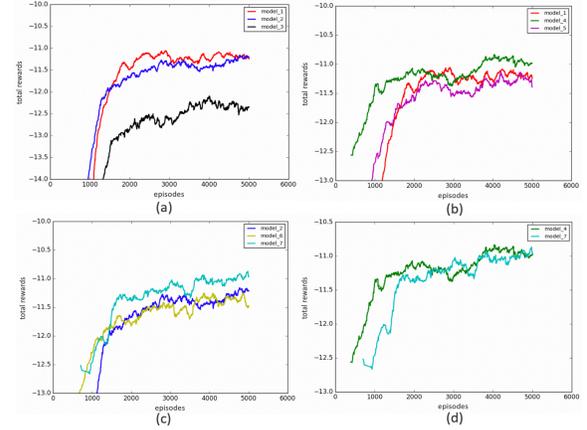

Figure 3. Total rewards accumulated during training (a) results of models ①, ②, ③; (b) results of models ①, ④, ⑤; (c) results of models ②, ⑥, ⑦; (d) results of models ④, ⑦.

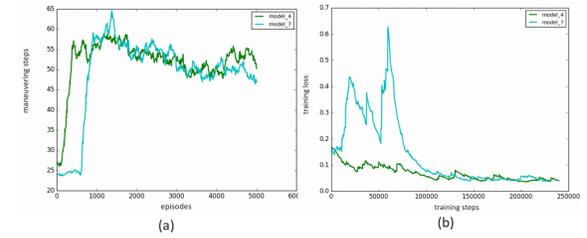

Figure. 4. Training curves of model models ④ and ⑦ (a) accumulated maneuvering steps in training episodes (b) training loss along training steps.

In addition, from Fig. 4(a) we can also learn that the RL vehicle agent picks up the lane change ability quite quickly as the maneuvering steps stay low just for a few hundreds of episodes and then rise rapidly. It can also be noticed that there is a small hump in between 1000 and 3000 episodes which means the lane change process lasts a little bit longer in that training period. This is consistent with the lane change scenarios that we will discuss in Fig. 5 where obvious swinging behavior is observed in episodes around 2000-3000 and then adjusted properly.

### 2) Driving performance progress during training

To analyze the progress in learning, we save a series of checkpoints (i.e. intermediate models) during training at episodes of [100, 200, 300, …, 4900, 5000]. Fig. 5 shows the visualization of some representative lane change scenes at different saved checkpoint, i.e. [500,

1000, 2000, …, 5000]. The red vehicle is the RL agent which intends to change to the target gap where the leading and lagging vehicles are marked as green. Other vehicles are marked as blue.

Fig. 5(a) demonstrates the lane-change behavior at episode of 500 where the RL agent always goes beyond the road borders (the dotted red line is an illustration of the past trajectory). That is because it learns from scratch and is ignorant of what the good actions are.

Fig. 5(b) shows the scenarios that the vehicle changes to the opposite direction, i.e. the selected gap is on the right lane but the vehicle changes to the left lane. The reason of this phenomenon is that the action space is randomly explored and there is some possibility of choosing an action that is opposite to the target gap especially when it has not gained sufficient knowledge of the rewarding rules.

Fig. 5(c) shows that the RL agent has learned some sense of the rewarding rules, at the episode of 2000. It knows not to go far beyond the border since the penalty will be quite large (see the design of equation (11)), so it swings around the boarder.

When training episode proceeds to 3000, the vehicle agent can make lane changes with just slightly swinging, as shown in Fig. 5(d). Gradually, the learning agent can behave like a novice driver.

It can change to the target lane without obviously running out of boundary but still needs adjusting the offset to stay at the center of the lane, as shown in Fig. 5(e).

Fig. 5(f) demonstrates successful lane change scenarios in which the RL agent can make either right or left lane changes smoothly with a single shot and without swinging or noticeable adjusting anymore. Video is available at https://youtu.be/UvMGwqfYY3k.

*3) Model validation with average performance*

A single vehicle's performance under a certain checkpoint (e.g. intermediate model) is not sufficient enough to claim that the RL agent can behave well generally and can adapt to diverse situations. Therefore, we select a set of checkpoints at episodes of [100, 400, 800, 1200, …, 4600, 5000] and run 100 lane-change cases for each saved checkpoint, and then calculate the average total returns at each checkpoint to check whether they converge the same way as the total rewards do in the training.

As shown in Fig. 6, it shows a very clear increasing and plateauing tendency as expected. At the beginning (checkpoint = 100), the RL agent cannot make lane changes in all simulation runs so that the average total return is low (equals the default value of -20). At around the checkpoint of 400-800, RL vehicle can make some acceptable lane change steps in some cases and the average return increases a lot. In the following checkpoints, the average returns stay roughly steady indicating that the RL agent is learning to adjust its behavior. The plateau in the latter part is also a good proof that the RL agent can always make stable lane changes in different situations since the total returns stay similar.

Another important evidence is the successful rate of the lane-change maneuver. Under the last saved checkpoint, there are 80 out of 100 cases in which the RL agent has performed lane change actions (other cases do not show lane change actions probably because there is no appropriate gap available), and in all of them the vehicle satisfactorily completes the task, with a 100% successful rate, even though it learns from scratch.

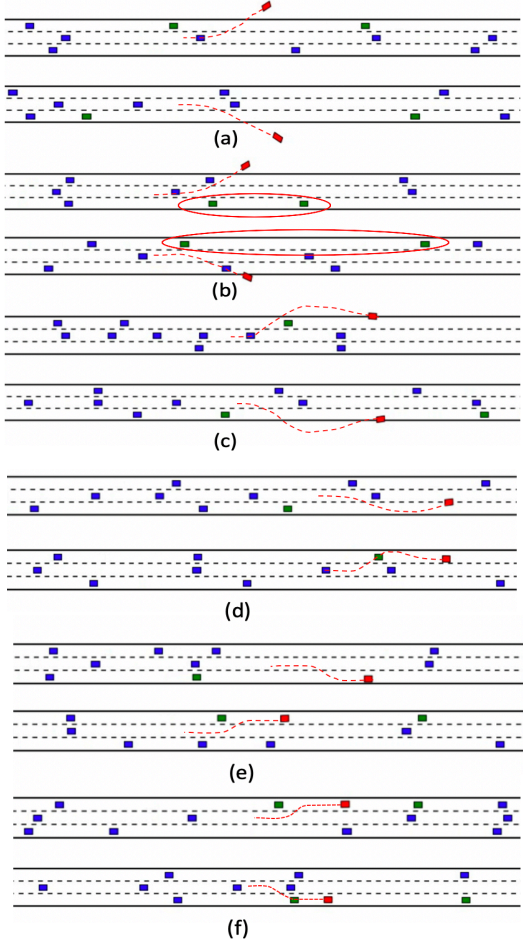

Figure 5. Scenario shots of representative training cases (red: ego vehicle, green: vehicles in the selected gap, blue: other vehicles, red curve: an illustration of past trajectory of the ego vehicle). (a) out of boundary, (b) lane changes to the opposite direction, (c) obvious swinging, (d) slightly swinging, (e) near successful lane changes, (f) successful lane changes.

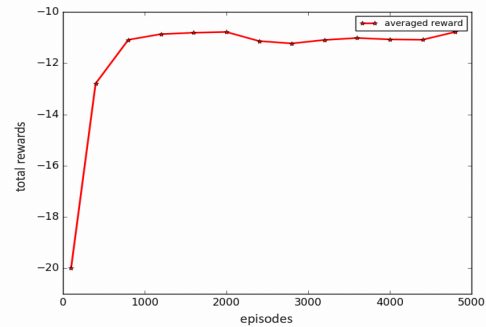

Figure 6. Average total returns under saved checkpoints

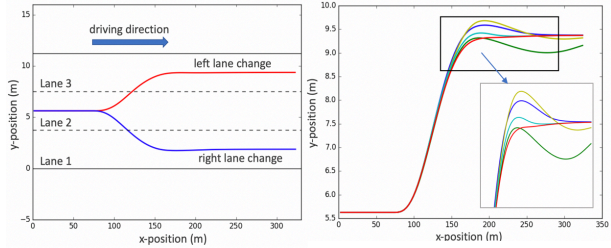

Figure 7. Lane-change trajectories under final checkpoint (a) trajectories for left and right lane changes, (b) example of gradually improved lane change behavior.

*4) Dynamics of learned lane change behavior*

Since we do not include a trajectory planning model in our system structure and the RL agent is designed to learn from scratch, it is important to know how the lane change trajectory looks like and how convincing the learned driving policy is. We plot the left and right lane-change trajectories of the model saved at the last checkpoint (episode = 5000).

As shown in Fig. 7(a), the lane change trajectories are quite smooth and stable. We also demonstrate the evolving progress of the left lane change trajectory, as in Fig. 7(b), as an illustration of how the agent adjusted its driving trajectory as learning evolves. The red curve in Fig. 7(b) is the learned trajectory from the final model and other colored curves are trajectories in the evolving progress.

## V. Conclusion and Discussion

In this work, we applied DDPG for learning the lateral movement of a lane change behavior from scratch. The action space is taken to be continuous. The reward function takes into account lateral control actions, position deviations and maneuvering time, and provide informative and guiding signals to the RL agent when it explores the continuous state and action space.

Simulation results show satisfactory learning curve where the RL vehicle agent initially always runs out of road boundary but eventually moves to the target lane smoothly and consistently with a successful rate of 100%. Validation results of averaged total returns also show expected convergence trend.

A potential application of the learned control policy in this study is in trajectory planning module. Since the policy is learned under dynamic driving environment, it naturally takes into account the changing environment and can generate rollouts based on the current state of the environment. The generated trajectory can be then tracked by a low-level controller.

It is also worth trying to use the RL model to learn both lateral and longitudinal control behavior simultaneously in one framework. Meanwhile, the gap selection function should also be incorporated into the learning procedure, making the overall formulation capable of handling full autonomy in autonomous driving maneuvers. This will be a central topic of our future studies.